\documentclass[lettersize,journal]{IEEEtran}
\IEEEoverridecommandlockouts
\usepackage{amsmath,amsfonts}
\usepackage{amssymb}
\usepackage{algorithmic}
\usepackage{algorithm}
\usepackage{array}
\usepackage[FIGTOPCAP]{subfigure}
\usepackage{textcomp}
\usepackage{stfloats}
\usepackage{url}
\usepackage{subcaption}

\usepackage{rotating}

\usepackage{verbatim}
\usepackage{graphicx}
\usepackage{cite}
\usepackage{xcolor}

\usepackage{multirow}
\usepackage{array}
\newcolumntype{P}[1]{>{\centering\arraybackslash}p{#1}}
\newcolumntype{M}[1]{>{\centering\arraybackslash}m{#1}}

\usepackage{hyperref}
\hypersetup{
    colorlinks=true,
    linkcolor=blue,
    filecolor=magenta,      
    urlcolor=blue,
    citecolor=black,
    pdftitle={Twilight SLAM},
    pdfpagemode=FullScreen,
    }

\def\BibTeX{{\rm B\kern-.05em{\sc i\kern-.025em b}\kern-.08em
    T\kern-.1667em\lower.7ex\hbox{E}\kern-.125emX}}

\begin{document}

\title{Twilight SLAM: Navigating Low-Light Environments}
\author{IEEE Publication Technology,~\IEEEmembership{Staff,~IEEE,}
\author{Surya Pratap Singh$^{1}$, Billy Mazotti$^{1*}$, Dhyey Manish Rajani$^{1*}$, Sarvesh Mayilvahanan$^{1}$, Guoyuan Li$^{2}$, Maani Ghaffari$^{1}$ \\ 
}
}


\maketitle

\begin{abstract}
This paper presents a detailed examination of low-light visual Simultaneous Localization and Mapping (SLAM) pipelines, focusing on the integration of state-of-the-art (SOTA) low-light image enhancement algorithms with standard and contemporary SLAM frameworks. The primary objective of our work is to address a pivotal question: Does illuminating visual input significantly improve localization accuracy in both semi-dark and dark environments? In contrast to previous works that primarily address partially dim-lit datasets, we comprehensively evaluate various low-light SLAM pipelines across obscurely-lit environments. Employing a meticulous experimental approach, we qualitatively and quantitatively assess different combinations of image enhancers and SLAM frameworks, identifying the best-performing combinations for feature-based visual SLAM. The findings advance low-light SLAM by highlighting the practical implications of enhancing visual input for improved localization accuracy in challenging lighting conditions. This paper also offers valuable insights, encouraging further exploration of visual enhancement strategies for enhanced SLAM performance in real-world scenarios.
\end{abstract}


\section{Introduction}

\footnotetext[1]{The authors are with the Department of Robotics, University of Michigan, 48109 Ann Arbor, USA}
\def\thefootnote{*}\footnotetext{Equal contribution; these authors can swap ordering as per their need}\def\thefootnote{\arabic{footnote}}
\footnotetext[2]{The authors are with the Department of Electrical \& Computer Engineering, University of Michigan, 48109 Ann Arbor, USA}
\footnotetext[3]{Corresponding e-mail: suryasin@umich.edu; Other authors email: drajani@umich.edu; bmazotti@umich.edu; smayil@umich.edu; lguoyuan@umich.edu; maanigj@umich.edu} 

\IEEEPARstart{I}{n} the realm of autonomous robotic systems, SLAM stands as a fundamental need, enabling robots and physical embodied agents to autonomously navigate and develop spatial awareness in unknown environments. These SLAM algorithms work by simultaneously estimating the agent’s location and constructing an environment map, ensuring a seamless and accurate exploration process.
The evolution of SLAM has seen a recent surge in interest, particularly in the domain of Visual SLAM (VSLAM). The appeal of Visual SLAM is fueled by the growing accessibility of low-cost sensors, their capability to capture rich environmental information, and their seamless integration with other sensor modalities \cite{karlsson2005vslam}. These attributes position VSLAM as an attractive choice for a diverse range of applications, spanning from robotics to augmented reality. Visual methods for camera pose estimation, a pivotal element of VSLAM, have garnered significant attention, proving their robust and sustainable functionality in both academic research and commercial applications \cite{savinykh2022darkslam}.
Visual SLAM relies solely on visual data for localization and mapping, unlike traditional methods with additional sensors like LiDAR.

While promising, Visual SLAM encounters challenges limiting its broad use, both indoors (e.g., hospitals, warehouses \cite{kalinov2020warevision}) and outdoors, necessitating adequate lighting and potentially raising operational costs in fully unmanned facilities.
\begin{figure}[h]
\centering
\includegraphics[width=7cm, height=7cm]{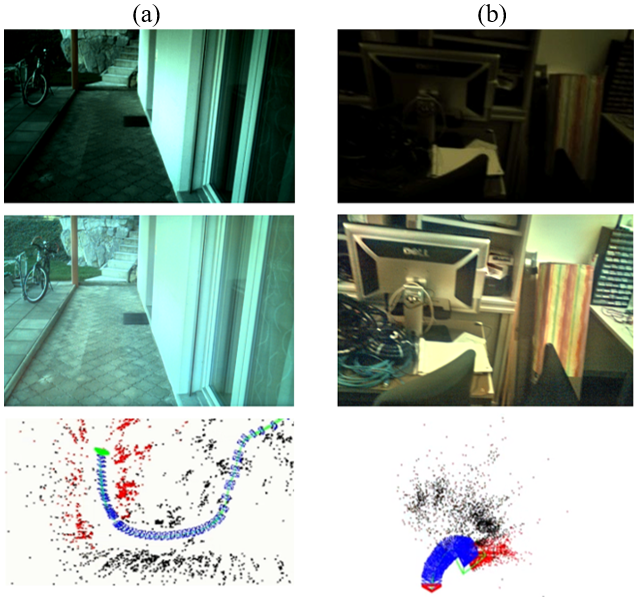}
\caption{Our customized low-light SLAM: (a) Semi-dark and (b) Dark scenes (Top) with Enhanced views (Middle) and Keyframe perspective (bottom)}
\label{fig:cover}  
\end{figure} 

One critical challenge impeding the efficacy of Visual SLAM algorithms is their robustness to varying illumination conditions, as they strongly depend on the quality of input images, leading to decreased localization accuracy in low light. The performance of feature extractors, a vital component of VSLAM, sharply declines in low-light environments, compromising the reliability and precision of the algorithms \cite{7782813}. Addressing these challenges has prompted various deep-learning-based approaches \cite{jiang2021enlightengan}, \cite{guo2020zero}. While these approaches have presented localization accuracies for their visual algorithms in "dim" conditions, they haven't explicitly carried out detailed analyses on semi-dark and challenging dark environments. So in our work, we establish threshold values for categorizing images based on normalized mean intensity i.e. dark [mean intensity$\leq$0.15], semi-dark 
[0.15$\leq$mean intensity$\leq$0.28] (shown in Fig. \ref{fig:cover}), and mild-daylight [0.28$\leq$mean intensity$\leq$0.35]. This selection, informed by statistical analysis and histogram examination, ensures a systematic framework that balances perceptual information and statistical characteristics in distinguishing illuminance levels.

\section{Related Work}
VSLAM formulations are broadly categorized into two groups: Feature-based (or direct) \cite{ferrera2021ov} and Image-dependent (or indirect) approaches \cite{min2021voldor+}. Feature-based methods use descriptors to identify images by their features, while Image-dependent approaches process entire images, often employing optical flow-based techniques. Each group, along with its sub-categorical formulations, has inherent limitations. Image-dependent approaches struggle with sudden changes in viewing perspective or angle, while feature-dependent frameworks are sensitive to lighting and contrast variations, impacting mapping and localization accuracy. Feature-based SLAM techniques, exemplified by works like \cite{ferrera2021ov}, \cite{mur2017orb}, and \cite{zhao2020good}, rely heavily on feature extractors, particularly key-points. However, in low-light environments, conventional descriptors, like SIFT, SURF, and FAST, lead to decreased localization accuracy.

Feature-based methods are constrained by the attributes of the descriptors they leverage. Unsymmetrical SIFT extractor exhibits superior illumination invariance, as noted by \cite{ross2013novel} and \cite{ross2014method}. Despite the prevalence of the ORB feature extractor \cite{rublee2011orb} in modern SLAM frameworks, it is overlooked in the aforementioned studies. Learning-based feature attribute description approaches like SuperGlue \cite{sarlin2020superglue} and SuperPoint \cite{detone2018superpoint} face limited adoption due to computational inefficiency and compatibility issues with standard algorithms. Several studies address appearance transformation, such as \cite{sarlin2021back}, which identifies the adaptability and invariance of features, though it lacks direct application in low-light scenarios. Another study \cite{8961635} utilizes basic image processing and warping for poorly-lit image sequences but is primarily tested in partially dim-light conditions, lacking robust evaluation in low-light environments. Research by \cite{porav2018adversarial} introduces GANs in the image pre-processing pipeline to enhance localization results. This study focuses on changing seasons and employs GAN-assisted unpaired image-to-image translation for accurate localization but has a low ratio of low-light to well-lit images. Upon further analysis, Savinykh et al. \cite{savinykh2022darkslam} propose DarkSLAM, a GAN-based SLAM method designed for low-light conditions. Their study highlights EnlightenGAN \cite{jiang2021enlightengan} as an effective image enhancer, but alternative SOTA methods like Zero-DCE \cite{guo2020zero} and Bread \cite{guo2023low} remain unexplored in SLAM and feature matching formulations.

To thoroughly examine various low-light image enhancement models and SLAM frameworks, we conduct an in-depth analysis with two main outcomes:
\begin{itemize}
\item Assess the systemic applicability of various SOTA image enhancement modules (like EnlightenGAN, Bread, Zero-DCE \& Dual\cite{zhang2019dual}) on challenging low-light conditions.
\item Evaluate a standard and a SOTA SLAM framework on relevant and unexplored low-light datasets, along with various image enhancement modules to draw conclusions about the best configurations to use.
\end{itemize}
The development of the above motivation and outcomes is described at length in the following sections.

\section{Datasets}
\subsection{KITTI}
KITTI (Karlsruhe Institute of Technology and Toyota Technological Institute) is one of the most extensively employed datasets in the domain of autonomous navigation. It encompasses several hours' worth of recorded traffic scenarios, utilizing diverse sensor modalities such as high-resolution RGB, grayscale stereo cameras, and 3D laser scanner. The dataset features image sequences with a resolution of 1,242×375, captured at 10 FPS (Frames per second) from an automobile navigating through urban and highway environments. In this paper, we specifically utilized sequences 04, 06, and 07. Notably, sequence 04 lacks a closed loop, while sequences 06 and 07 incorporate closed loops. While our primary emphasis is on enhancing low-light scenes, these mild-daylight image sequences also provide valuable insights into the adaptability and effectiveness of image enhancers in various environmental lighting scenarios.
\subsection{ETH3D}

ETH Zurich's ETH3D dataset focusing on 3D vision and robotics, has contributed significantly to SLAM research by releasing datasets with varying illumination levels, accessible on their website. For our study, we leverage two of their image sequences, namely \textit{sfm\_house\_loop} and \textit{sfm\_lab\_room2}, which were specifically collected using a handheld stereo camera in both indoor and outdoor settings. Each sub-dataset within this collection features a raw image sequence, with or without loop closure, and segregated data in monocular, stereo, RGB-D, and IMU formats. These formats encompass calibration, ground truth, and RGB images constituting the image sequence. The \textit{sfm\_house\_loop} dataset captures a camera moving in a large loop around a house, while the \textit{sfm\_lab\_room2} dataset documents a camera navigating through a cluttered lab. Both datasets were filmed without the availability of a Vicon mocap system, necessitating the determination of ground truth using Structure-from-Motion.

Our selection of the \textit{sfm\_house\_loop} and \textit{sfm\_lab\_room2} datasets is primarily motivated by their extensive coverage of traversing in poorly-lit, and dark environments, respectively. This choice proves pivotal in configuring the framework for the objectives of our study.

\section{Image Enhancement Modules}
\label{IV}

An integral component of our model pipeline is the image enhancement module, responsible for illuminating low-light images for integration into the SLAM methods mentioned in Section \ref{V}. In this study, we conducted rigorous validation of four distinct SOTA enlightening models. These models were carefully selected based on their performance on conventional low-light image enhancement datasets. Subsequently, they were incorporated into our low-light SLAM pipelines.

\subsection{EnlightenGAN}

EnlightenGAN \cite{jiang2021enlightengan} is an unsupervised generative adversarial network trained without a low/normal-light image pair dataset. Unlike many SOTA deep learning models relying on synthetic or highly cleaned datasets with ground-truth enlightened images, such data's availability and quality are insufficient for real-world applications.

The training process eliminates the need for ground truth enlightened images by establishing an unpaired mapping between low/normal-light images without relying on perfectly paired images. This is achieved through a global and local discriminator. The global discriminator compares the model's outputted enlightened image to another relatively similar normal light image. Simultaneously, the local discriminator samples random cropped sections of the enlightened and normal light images, comparing the sections. If the model can't discriminate between the two sections, the loss is low, and vice versa. The use of global and local discriminators ensures EnlightenGAN's good performance in smaller regions as well as the image as a whole. Due to the robustness of this unpaired training, EnlightenGAN excels in enhancing real-world images across various domains, making it highly applicable for general-purpose low-light SLAM.

\begin{figure}[htp] %
\includegraphics[width=9cm, height=7.5cm]{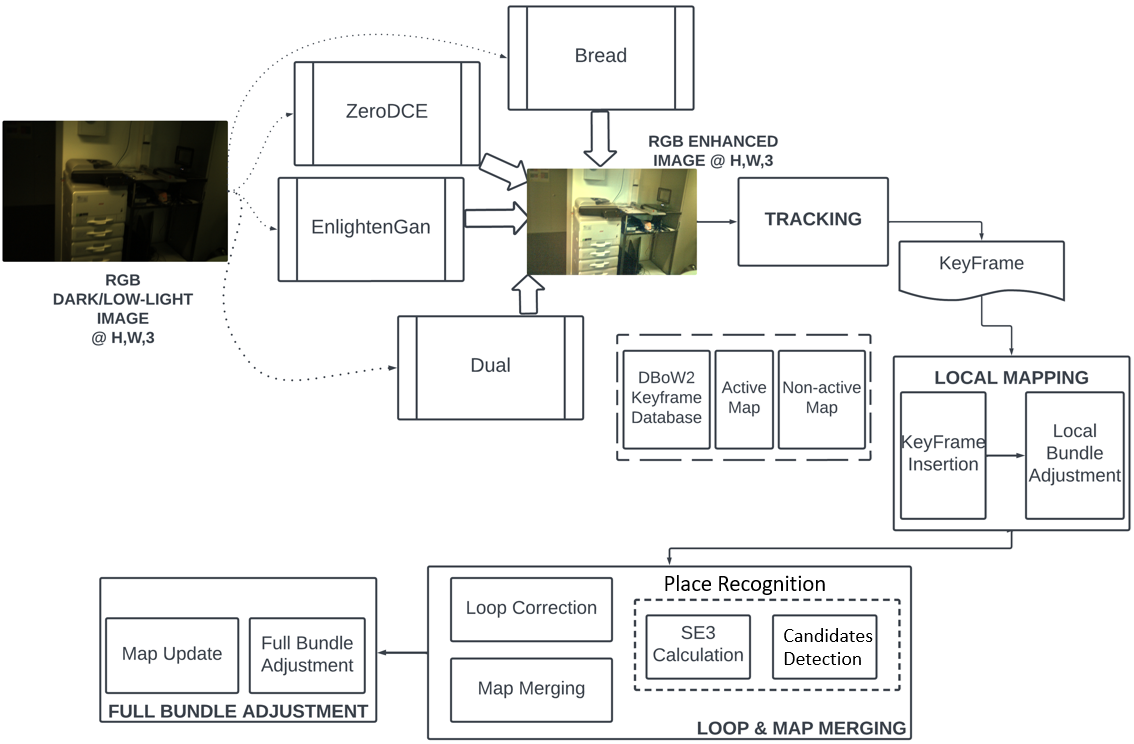}
\caption{Customized ORB-SLAM3 Pipeline}
\label{fig:orbslam3}  
\end{figure}

\subsection{Bread}

Bread \cite{guo2023low} introduces a novel low-light enhancement model that uniquely addresses color and noise enhancement separately. This framework highlights a significant advancement achieved by decoupling noise and color through the conversion of the image from RGB to luminance (brightness) and chrominance (color) spaces.
The process begins by using an Illumination Adjustment Network to brighten the image, then passing it through a noise suppression module to eliminate noise in the brightened luminance. The enhanced luminance undergoes further refinement in the Color Adaptation Network, leveraging the illumination map to generate realistic colors both globally and locally.
Comparing Bread to other models on the Low-Light dataset, it outperforms in all metrics, including peak signal-to-noise ratio (PSNR) and structural similarity index measure (SSIM). This showcases Bread's excellence in addressing low-light scenarios and its comprehensive approach to color and noise enhancement.

\subsection{Zero-DCE}

Zero-Reference Deep Curve Estimation (Zero-DCE) \cite{guo2020zero} is a novel deep-learning-based low-light image enhancement method, training a lightweight deep network to estimate pixel-wise and high-order curves for dynamic range adjustment of given images. Unlike traditional image-to-image translation-based enhancement methods, Zero-DCE reformulates the task as an image-specific curve estimation problem. It takes low-light images as inputs into a Deep Curve Estimation Network (DCE-Net) and estimates a set of high-order Light-Enhancement curves (L-E curves) as its output. The non-reference loss function performs pixel-wise adjustment on the dynamic range of the input images' RGB channels by applying the high-order curves, ultimately obtaining an enhanced image.

This method offers unique advantages, making it a robust and desirable image enhancement technique for low-light visual SLAM projects. The training process is more convenient than CNN and GAN-based methods, as it omits the need for paired and unpaired reference data, respectively. Additionally, it adapts well to various lighting settings, including nonuniform and poor light cases, brightening low-light images without compromising their inherent features or quality. Finally, it proves computationally effective and efficient.

\subsection{Dual}

Dual Illumination Estimation for Robust Exposure Correction \cite{zhang2019dual} is an innovative automatic exposure correction method utilized in this study as an image-enhancement technique. The framework initiates with dual illumination estimation on both the original and inverted input low-light images, extracting forward and reverse illuminations. Subsequently, it recovers intermediate under-exposure and over-exposure corrected images by solving related objective functions, utilizing well-tuned correction parameters and a refined illumination map. An effective multi-exposure image fusion follows, blending visually best-exposed parts from the two intermediate images and the input image, utilizing well-tuned influence parameters to create a globally well-exposed final image.

Notably, this method boasts a simple yet effective framework, capable of producing high-quality exposure-corrected results for images with diverse exposure conditions, including underexposed, overexposed, partially under-exposed, and partially over-exposed cases. Due to its versatility and reliability, it is deemed an ideal low-light image enhancement technique for integration into our low-light visual SLAM pipelines.

\begin{figure*}
\centering
\includegraphics[width=18cm, height=7cm]{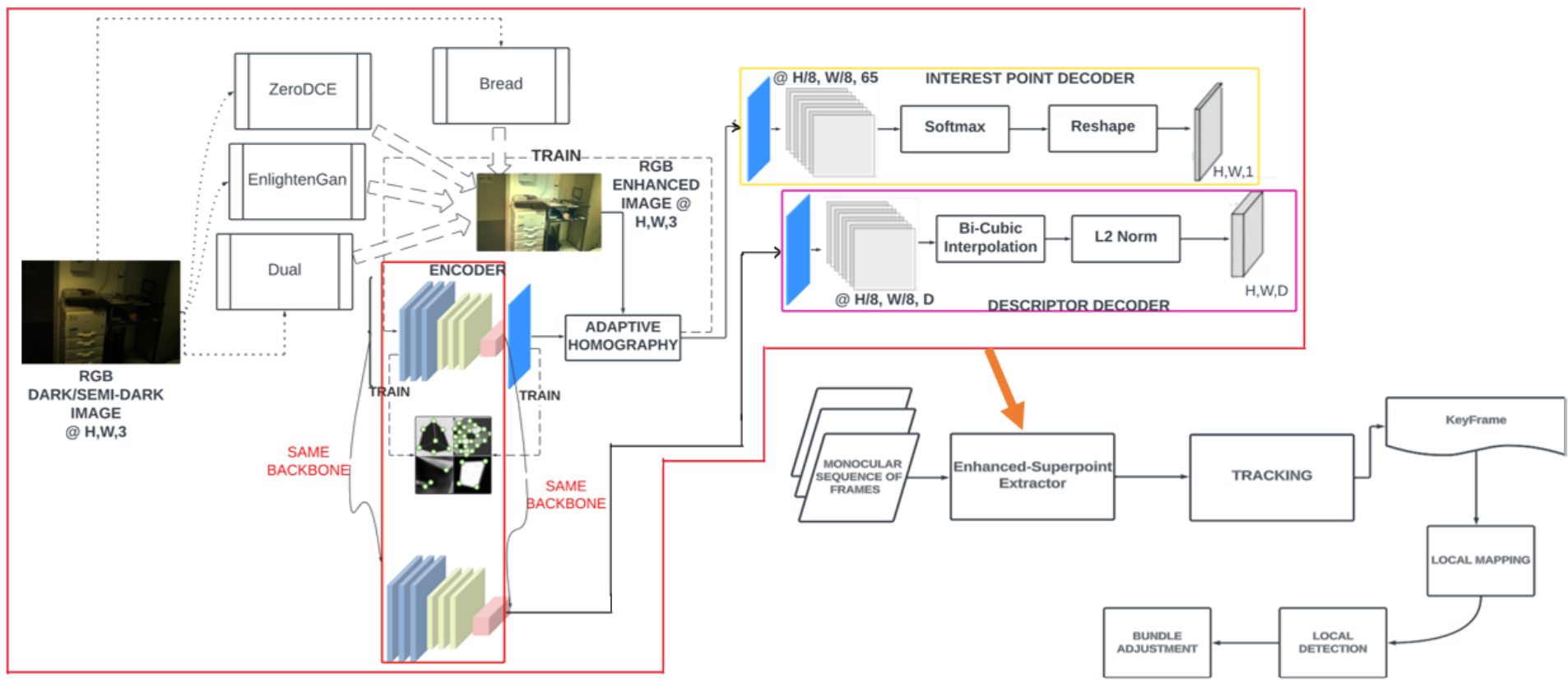}

\caption{Customized SuperPoint-SLAM Pipeline}
\label{fig:superpoint}  
\end{figure*}

\subsection{Comparison}

\begin{figure*}
\centering
\includegraphics[width=16cm, height=4cm]{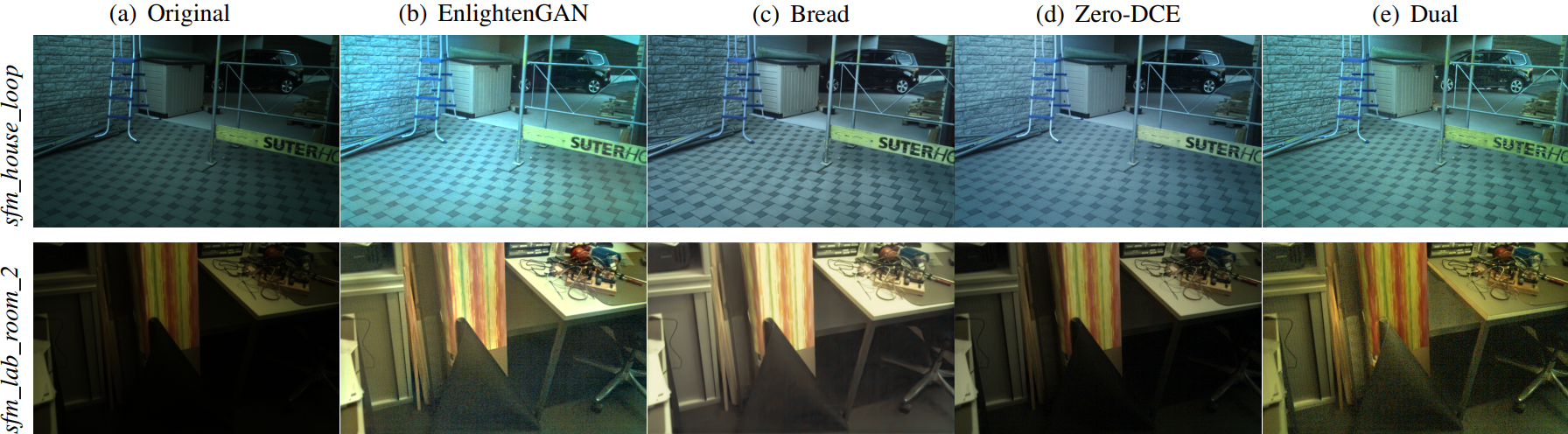}
\caption{Comparing image enhancement modules for two different ETH3D low-light SLAM datasets}
\label{fig:image_enh_comparision}  
\end{figure*}

Figure \ref{fig:image_enh_comparision} presents a comparison of the outputs from various image enhancement modules across select frames from two distinct low-light datasets. A qualitative assessment of their performance reveals that EnlightenGAN, Bread, and Dual exhibit effective illumination of the images. However, Zero-DCE seems to encounter difficulties, especially evident in its struggles with the dark indoor images of \textit{sfm\_lab\_room2}.

\section{SLAM Methods}
\label{V}
To broaden the scope of this study, we assess the effects of the image enhancement modules detailed in Section \ref{IV} on two standard SLAM frameworks. Both ORB-SLAM3 and SuperPoint-SLAM, while building on the ORB-SLAM architecture, employ deterministic and learning-based features, respectively. The inherent distinctions in the feature detectors and descriptors between these two SLAM frameworks enable us to draw empirical-based conclusions regarding the impacts of different image enhancement modules on SOTA feature detectors and descriptors for SLAM applications.

\subsection{ORB-SLAM3}

ORB-SLAM3, a visual sensor-based SLAM architecture designed for Monocular, Stereo, and RGB-D cameras\cite{campos2021orb}, stands out for its adaptability and high precision. It has the capability to compute the camera trajectory and a sparse 3D scene reconstruction in real-time, catering to scenarios ranging from brief hand-held sequences to city block driving\cite{mur2017orb}. Evolving from the foundations of ORB-SLAM2 and ORBSLAM-VI\cite{theodorou2022visual}, ORB-SLAM3 emerges as a comprehensive multi-map and multi-session system. It operates in either visual-inertial or pure visual modes, supporting both pinhole and fisheye lens models. Figure \ref{fig:orbslam3} illustrates the components of our ORB-SLAM3 pipeline.

In our research study, we chose ORB-SLAM3 over ORB-SLAM2 due to significant novelties addressing limitations in the latter. Notably, ORB-SLAM3 introduces a multi-map system with an enhanced place recognition technique, improving recall. This modification allows ORB-SLAM3 to operate effectively even during extended periods of limited visual information. In instances where the agent loses its position, it initiates a new map, seamlessly integrating it with previously generated maps upon revisiting those areas\cite{campos2021orb}.

\subsection{SuperPoint-SLAM}
SuperPoint-SLAM is built upon the ORB-SLAM2 architecture, replacing ORB-SLAM2's deterministic feature detector and descriptor, ORB, with SuperPoint, a learning-based feature detector and descriptor. The training of SuperPoint involves using a pre-trained interest point detector and Homographic Adaptation to generate pseudo-ground truth interest points in real images, as depicted in Figure \ref{fig:superpoint}.


\begin{table*}[h]
    \centering
    \captionsetup{justification=centering}
    \caption{NCPP, RMSE, and max error for all SLAM pipelines \\ (red = best performance for SLAM framework, blue = best performance overall)}
    \begin{tabular}{ |M{1.2cm}|M{1.8cm}|M{1cm}M{0.8cm}M{0.75cm}M{0.75cm}M{1.2cm}|M{1cm}M{0.8cm}M{0.75cm}M{0.75cm}M{1.2cm}|  }
        \hline
        \multicolumn{2}{|c|}{} & \multicolumn{5}{c|}{Superpoint-SLAM} & \multicolumn{5}{c|}{ORB-SLAM3} \\
        \hline
        Sequence & Metric & Original & Bread & Dual & EnGAN & Zero-DCE & Original & Bread & Dual & EnGAN & Zero-DCE \\
        \hline
        \multirow{3}{*}{KITTI 04} & NCPP (pairs) $\uparrow$ & 243 & 223 & 236 & \textcolor{blue}{\textbf{244}} & 241 & 159 & 188 & \textcolor{red}{\textbf{205}} & 192 & 167 \\
        & RMSE (m) $\downarrow$ & \textcolor{blue}{\textbf{0.26}} & 0.40 & 0.47 & 0.37 & 0.52 & 1.72 & 1.30 & 1.52 & \textcolor{red}{\textbf{1.19}} & 1.50 \\
        & Max AE (m) $\downarrow$ & \textcolor{blue}{\textbf{0.50}} & 0.80 & 1.08 & 0.77 & 1.13 & 3.59 & 3.01 & 3.27 & \textcolor{red}{\textbf{2.71}} & 3.85\\ \hline
        \multirow{3}{*}{KITTI 06} & NCPP (pairs) $\uparrow$ & 603 & 412 & 589 & 586 & \textcolor{red}{\textbf{604}} & 575 & 618 & \textcolor{blue}{\textbf{717}} & 625 & 607 \\
        & RMSE (m) $\downarrow$ & 14.73 & \textcolor{blue}{\textbf{11.00}} & 13.03 & 14.50 & 14.12 & 17.65 & 14.56 & 13.91 & 15.22 & \textcolor{red}{\textbf{11.98}} \\
        & Max AE (m) $\downarrow$ & 23.68 & 22.13 & \textcolor{red}{\textbf{21.70}} & 22.90 & 27.22 & 35.43 & 28.49 & 24.24 & 22.79 & \textcolor{blue}{\textbf{18.39}}\\ \hline
        \multirow{3}{*}{KITTI 07} & NCPP (pairs) $\uparrow$ & 555 & 564 & 568 & 552 & \textcolor{red}{\textbf{598}} & 589 & 613 & \textcolor{blue}{\textbf{721}} & 615 & 603\\
        & RMSE (m) $\downarrow$ & 2.40 & 2.57 & 8.07 & 3.22 & \textcolor{red}{\textbf{2.08}} & 2.49 & 2.85 & 4.85 & \textcolor{blue}{\textbf{1.81}} & 2.14 \\
        & Max AE (m) $\downarrow$ & 5.53 & 6.90 & 21.48 & 6.61 & \textcolor{red}{\textbf{4.51}} & 4.79 & 4.97 & 8.01 & \textcolor{blue}{\textbf{3.36}} & 3.65 \\ \hline
        \multirow{3}{*}{sfm\_house} & NCPP (pairs) $\uparrow$ &  27 & 22 & 24 & 29 & \textcolor{red}{\textbf{34}} & \textcolor{blue}{\textbf{313}} & 289 & 290 & 281 & 285\\
        & RMSE (m) $\downarrow$ & \textcolor{red}{\textbf{0.23}} & 0.25 & 0.45 & 0.57 & 0.67 & 0.43 & 0.29 & 0.54 & \textcolor{blue}{\textbf{0.15}} & 0.36 \\
        & Max AE (m) $\downarrow$ & \textcolor{blue}{\textbf{0.44}} & 0.47 & 0.49 & 0.88 & 1.01 & 0.96 & 0.92 & 1.22 & \textcolor{red}{\textbf{0.46}} & 0.95\\ \hline
        \multirow{3}{*}{sfm\_lab} & NCPP (pairs) $\uparrow$ & 11 & \textcolor{red}{\textbf{13}} & \textcolor{red}{\textbf{13}} & 12 & 10 & 63 & 88 & \textcolor{blue}{\textbf{96}} & 90 & 90\\
        & RMSE (m) $\downarrow$ &7.28 & 7.27 & 7.22 & \textcolor{red}{\textbf{7.21}} & 7.26 & 1.13 & 0.53 & 0.53 & \textcolor{blue}{\textbf{0.52}} & 0.56\\
        & Max AE (m) $\downarrow$ & 7.29 & 7.29 & 7.26 & \textcolor{red}{\textbf{7.23}} & 7.28 & 1.51 & 0.87 & 1.03 & \textcolor{blue}{\textbf{0.78}} & 0.96\\
        \hline
    \end{tabular}
    \label{table:results}
\end{table*}





The integration of SuperPoint, along with the SuperGlue matching algorithm, yields competitive performance in Visual Place Recognition and Image Matching tasks on Berlin Kudamm and MC PhotoTourism benchmarks\cite{sarlin2020superglue}. Furthermore, SuperPoint's computationally efficient architecture achieves real-time processing at 70 FPS with a Titan X GPU \cite{detone2018superpoint}.

Extending on SuperPoint's widespread adoption with popular feature extractors like ORB and GCNv2, the authors of SuperPoint-SLAM investigated variations in ORB-SLAM performance by substituting each feature detector along with its associated binary bag of words, following the approach proposed by \cite{galvez2012bags}. SuperPoint-SLAM, as reported by its authors, achieved competitive performances with ORB-SLAM on the KITTI SLAM benchmark dataset but encountered challenges in sequences with more cluttered off-road scenes. Beyond SuperPoint's feature equivariance and its extensive application in visual odometry tasks, SuperPoint-SLAM establishes itself as a robust foundation for SLAM, delivering effective performance in dynamic environments.

\section{Results}

In Table \ref{table:results}, we present a comparison of the impact of various image enhancement modules on KITTI sequences and ETH3D datasets using the monocular Superpoint-SLAM and ORB-SLAM3 architectures. The primary metric for comparison is the number of compared pose pairs (NCPP), indicating the features extracted throughout the sequence. Typically, low-light images yield fewer features, and image enhancement modules are expected to increase the number of extracted features. Additionally, we use two evaluation metrics, root mean square error (RMSE) and maximum absolute error (Max AE), to assess the discrepancy between SLAM-predicted positions and ground truth.

The mild-daylight dataset comprises KITTI 04, 06, and 07 sequences, the semi-dark dataset is the \textit{sfm\_house} sequence, and the dark dataset is the \textit{sfm\_lab} sequence. Due to constraints in computing resources, SuperPoint-SLAM required a notably longer duration compared to ORB-SLAM3 for processing the same dataset sequences. The combinations of image enhancers with ORB-SLAM3 and SuperPoint-SLAM encountered feature detection failures in some runs over ETH3D's semi-dark and dark datasets. This approach demonstrated superior performance across multiple metrics, including NCPP, RMSE, and MAE. Table \ref{table:results} presents best results from five runs for each sequence, prioritizing the lowest average RMSE before the highest NCPP. Furthermore, it produced trajectory maps that are qualitatively more accurate and expressive.

Analyzing the features extracted (using NCPP), all enhancement modules resulted in more features (with marginal improvement for SuperPoint-SLAM) compared to the original dataset, except for the semi-dark sequence (\textit{sfm\_house}) in ORB-SLAM3. Notably, the image enhancement modules enabled ORB-SLAM3 to achieve the lowest RMSE and Max AE for all dataset sequences. In contrast, for SuperPoint-SLAM, the modules improved RMSE and Max AE only for KITTI 06, \textit{sfm\_house}, and \textit{sfm\_lab} sequences.

Comparing overall performance, EnlightenGAN demonstrated the best results, as summarized in Table \ref{table:results-best-combo}, which showcases the best combinations of SLAM architectures and image enhancement modules under different lighting conditions.

\begin{table}[h]
    \centering
    \caption{Best Overall Combinations}
    \begin{tabular}{|M{1.8cm}|M{2cm}|M{2.6cm}|}
        \hline
        Dataset Type & ORB-SLAM3 & Superpoint-SLAM \\
        \hline
        mild-daylight & EnlightenGAN & Original \\
        \hline
        Semi-Dark & EnlightenGAN & Original \\
        \hline
        Dark & EnlightenGAN & EnlightenGAN \\
        \hline
    \end{tabular}
    \label{table:results-best-combo}
\end{table}

\begin{figure*}
\centering
\includegraphics[width=18cm, height=4.5cm, width=1.0\textwidth]{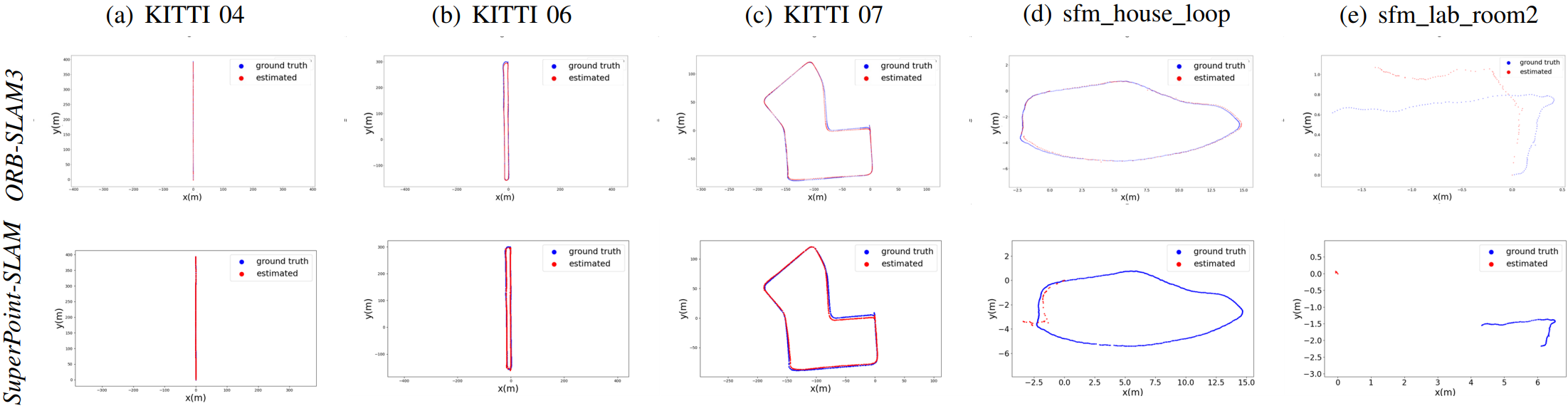}
\captionsetup{justification=centering}
\caption{Optimal combinations of SLAM architectures and image enhancement modules on KITTI and ETH3D SLAM datasets}
\label{fig:results_slam}  
\end{figure*}

In Figure \ref{fig:results_slam}, we present the trajectory maps of the best-run scenarios for ORB-SLAM3 and SuperPoint-SLAM. Examining these plots, we first focus on sequences with loop-closures, namely KITTI sequences 06 and 07, and the \textit{sfm\_house} loop. In these cases, we observe that the SLAM-generated poses closely align with the ground truth, indicating robust performance. However, shifting attention to KITTI sequence 04 and \textit{sfm\_lab} sequence, which lack loop-closures, we note that the SLAM-generated poses exhibit close alignment with the ground truth only in mild-daylight environments. Moreover, SuperPoint-SLAM performs poorly under extremely dark environments, even with image enhancers.

\section{Conclusion}

In this work, we conducted a comprehensive exploration of low-light visual SLAM pipelines, assessing a variety of combinations involving image enhancers and SLAM architectures across navigation datasets. Employing diverse metrics, we gauged the effectiveness of these combinations under varying lighting conditions. Furthermore, we delved into the strengths and weaknesses of each configuration, aiming to identify optimal settings for individual datasets.

Our findings demonstrate that enhancing visual input significantly improves localization accuracy in semi-dark environments, with noticeable enhancement observed in mild-daylight conditions. However, in extremely dark environments, the improvements in localization accuracy with SOTA enhancers are incremental. Challenges in dark environments, including feature detection failures, contribute to this limitation. 

Looking ahead, future research could investigate the phenomenon of why the ORB feature detector extracts more features from extremely dark enhanced image sequences compared to semi-dark image sequences. Exploring illumination map interpolation and overall integration of alternative enhancement estimation models like LIME \cite{7782813} with SOTA SLAM architectures, holds promise for further advancements in addressing the low-light navigational complexities.

\vspace{-0.15cm}
\section*{Supplementary material}
The code is publicly available at \url{https://github.com/TwilightSLAM}
\vspace{-0.15cm}
\bibliographystyle{IEEEtran}
\bibliography{IEEEabrv,refs.bib}

\end{document}